\documentclass[11pt,a4paper]{scrartcl} 
\usepackage[nochapters,eulermath]{classicthesis} 
\pdfoutput=1

\usepackage{lipsum}
\usepackage{graphicx}
\usepackage{tabularx}
\usepackage[left=1.5in,right=1.5in,top=1in,bottom=1in]{geometry}
\usepackage[nodoi]{apacite}

\usepackage{censor,caption}

\usepackage{listings}
\usepackage{fancyvrb}
\usepackage{pdflscape}
\usepackage{float}
\usepackage{adjustbox}
\usepackage{amsmath}
\usepackage{metrix}
\usepackage{enumerate}
\usepackage{covington} 
\usepackage[super]{nth} 

\usepackage[utf8]{inputenc}

\newenvironment{myverb}{%
 \VerbatimEnvironment
 \begin{adjustbox}{max width=\linewidth}
 \begin{BVerbatim}
  }{
  \end{BVerbatim}
 \end{adjustbox}
}

\usepackage{etoolbox}
\usepackage{setspace} 
\AtBeginEnvironment{quote}{\singlespace\vspace{0.5em}\small}
\AtEndEnvironment{quote}{\vspace{0.5em}\endsinglespace}

\begin{document}
\title{\rmfamily\normalfont\textsc{Metre as a stylometric feature in Latin hexameter poetry}}
\author{\phantom{xxx}\\\textsc{Ben Nagy}\\\small{\texttt{benjamin.nagy@adelaide.edu.au}}}
\date{\normalsize{\today}}

\maketitle

\begin{abstract}
\noindent
This paper demonstrates that metre is a privileged indicator of authorial style in classical Latin hexameter poetry. Using only metrical features, pairwise classification experiments are performed between 5 first-century authors (10 comparisons) using four different machine-learning models. The results showed a two-label classification accuracy of at least 95\% with samples as small as ten lines and no greater than eighty lines (up to around 500 words). These sample sizes are an order of magnitude smaller than those typically recommended for BOW (`bag of words') or $n$-gram approaches, and the reported accuracy is outstanding. Additionally, this paper explores the potential for novelty (forgery) detection, or `one-class classification'. An analysis of the disputed Aldine Additamentum (Sil. Ital. \textit{Puni.} 8:144-225) concludes ($p$=0.0013) that the metrical style differs significantly from that of the rest of the poem.\\
\phantom{xxx}
\end{abstract}

\setlength{\parindent}{0.3in}\section{Introduction}
While authorship attribution studies, and stylometry more generally, are becoming much more common, the bulk of the research has been applied to prose. In terms of work that considers poetry \textit{qua} poetry, the pickings are rather slim. \citeA{kao_computational_2012} attempt to computationally assess aesthetics. \citeA{omer_arud_2017} apply the Arabic science of poetic metre (\textit{Arud}) to authorship attribution, but the texts they analyse are all prose. Historically, some significant work has been done on Shakespeare but, for example, the recent computational results of \citeA{ilsemann_christopher_2018} have been sharply criticised by \citeA{barber_marlowe_2019} as a misapplication of Burrows' Delta \cite{burrows_delta:_2002}. \shortciteA{tizhoosh_poetic_2008} and \shortciteA{chaudhuri_small_nodate} both set about distinguishing poetry from prose, but offer no analysis of the poetry itself. One notable exception is \shortciteA{neidorf_large-scale_2019}, which sets out to demonstrate the single-authorship of Beowulf using a range of techniques, including a consideration of metre, albeit brief. This general lacuna has been noted by \shortciteA{plechac_versification_2019}, who have begun to treat metrical features as a first-class citizen of the stylometry universe, and are some of the few researchers applying the most recent multivariate and machine-learning techniques. One final paper that deserves attention is \shortciteA{forstall_evidence_2011}, which examines metre, specifically in Latin poetry. Unfortunately, the methods used by this team treated syllable quantities in Latin simplistically as an $n$-gram (a string of $n$ symbols, where each symbol may only be \metricsymbols{u} or \metricsymbols{_}), and they were unable to achieve useful separation between authors on this basis. One reason that poetry-specific approaches are important is that the best-known attribution methods for prose,%
\footnote{A good survey is available in \citeA{jockers_comparative_2010}. The present consensus seems to favour analysis based on frequency of closed-class words (eg function words) or character $n$-grams. However Bag of Words (BOW) approaches are still popular, in particular as implemented in the R stylo \texttt{rolling\_delta()} method \shortcite{r_stylo}.}
as aptly pointed out by \citeA{eder_does_2015}, require fairly large sample sizes, which are often not available for verse.%
\footnote{\citeA{eder_does_2015} suggests a 5000 word minimum for authorship studies on prose. Later \citeyear[4]{eder_short_2017} he revises this to ``as little as 2000 words in many cases'', which is still a great deal too many to be viable for many questions concerning poetry.\label{edernote}}
Although it does not rhyme, Latin hexameter offers a particularly rich interplay of metrical features, and it was hypothesised that the density of stylistic information contained in the poems' metre would allow for effective analysis with much less text.

\section{Research Questions}

I address two research questions in relation to the analysis of Latin hexameter poetry. The first, and most fundamental is this: ``Are the metrical attributes of Latin hexameter poetry an effective feature for stylometric analysis?''. I demonstrate, via two-label classification, that the answer is emphatically yes. Having validated the approach, I apply metrical analysis to the disputed authorship of the Aldine additamentum (Sil. Ital. \textit{Puni.} 8:144-225), seeking to re-address the question: ``Is this passage genuine?''.

\section{Methods}{
The website Pedecerto \shortcite{colombi_pedecerto_2007} has, for some time, offered on-demand, and highly accurate, scansion, particularly optimised for hexameters and elegaic couplets. In partnership with the site Musisque Deoque \cite{mqdq_2007}, they also maintain an extensive digital library of Latin poetry. This scansion is presented via their webpage as an image, showing the syllable quantities and other metrical information (caesurae, elision, hiatus, diaresis etc). In March, 2019, however, in response to my enquiry, their technical team added a new feature which allowed the download of any of the MQDQ poems in XML format, including the full scansion.%
\footnote{http://www.pedecerto.eu/pagine/autori, Retrieved \today.}
This offers a valuable resource to Digital Humanities practitioners, and made much of this present work possible. In Fig. \ref{fig:rawxml}, we can see the wealth of information available for each hexameter line. As well as the pattern of dactyls and spondees in the first four feet (\texttt{DDDS}) we can see the caesurae (\texttt{CM} or \texttt{CF} for strong and weak, respectively) and, most importantly, the full scansion of every syllable. Each syllable in the line is marked by two symbols, showing the foot in which it falls and a letter representing the position in that foot---so, for example, \texttt{1A1b1c} corresponding to \textit{vitaque} describes a three syllable word, with the \textit{arsis} (initial syllable) of the first foot followed by two \textit{breves} (short syllables).
\begin{figure}
\centering
\setlength{\belowcaptionskip}{10pt}
\caption{Raw XML output from MQDQ}
\label{fig:rawxml}
\begin{myverb}[]
<line name="952" metre="H" pattern="DDDS">
    <word sy="1A1b1c" wb="DI">Vitaque</word>
    <word sy="2A" wb="CM">cum</word>
    <word sy="2b2c3A" wb="CM">gemitu</word>
    <word sy="3b3c" wb="DI">fugit</word>
    <word sy="4A4T5A5b" wb="CF">indignata</word>
    <word sy="5c" wb="DI">sub</word>
    <word sy="6A6X">umbras.</word>
</line>
\end{myverb}
\end{figure}
\subsection{Feature Selection}
There are four general types of metrical features that were extracted from each line of hexameter (a concise list is available in the Appendix, Table \ref{fig:feat_abbrevs}). The first, which has been most often studied, is the metrical pattern in the first four feet. A hexameter line is constructed from six feet, each of which can be either a spondee (\metricsymbols{_ _}) or a dactyl (\metricsymbols{_ u u}). However by convention, the final two feet are almost always dactyl, spondee (\metricsymbols{_ u u , _ _}).%
\footnote{A spondee may appear in the fifth foot, but these are very rare. There are twenty-three such lines in the \textit{Aeneid} of Vergil, comprising around 0.2\%.}
This yields sixteen possible foot combinations. These combinations have been extensively studied by Classicists, the `recent' standard works being \citeA{platnauer_1951} and a string of publications by Duckworth in the 1960s.%
\footnote{cf, at the least, Duckworth \citeyearNP{duckworth_horaces_1965}, \citeyearNP{duckworth_five_1967}, \citeyearNP{duckworth_maphaeus_1969} \citeyearNP{duckworth_vergil_1969}. There are probably others.}
The next group of features are the \textit{caesurae}. A caesura occurs whenever there is a word break in the middle of a foot. Caesurae can, thus, occur after the initial spondee, which I will refer to as a \textit{strong caesura} or after the first breve in a dactyl, a \textit{weak caesura}.%
\footnote{An interested reader will find a range of other names in the literature: from the distasteful `masculine' vs `feminine', to several delightfully obscure technical terms, derived from the Greek, that encode both the strength and the position---\textit{pentethemimeral, hepthemimeral} etc.}
By convention, although any foot may (technically) contain a caesura, attention is restricted to the ones in the second, third and fourth feet. I should note that, in traditional analysis, a line of hexameter is considered to have a \textit{principal} caesura, which entails a pause; often a sense-pause, although sometimes simply a pause for breath. The present analysis makes no attempt to determine which, if any, of the available caesurae are 'principal'. It is important to understand, however, that the general `rule' was to place a strong caesura in the third foot, although every author deviated from this rule to a certain extent as a matter of stylistic preference.

The third class of features requires a little explanation. The first syllable of each of the six feet is called an \textit{ictus syllable}. This term, although Latin (lit. stroke, thrust, blow) derives from ancient Greek hexameter. Due to the language difference, however, it was not the ictus syllables in Latin verse that carried the primary stress---instead the Latin words were stressed (or accented) as usual. This interplay between Latin word accent and metrical ictus creates an inherent \textit{conflict}, which formed an important part of the style of Latin hexameter. Perhaps counter-intuitively, the general trend was to create conflict throughout the first four feet (otherwise, presumably, the line sounded too `sing-songy'), and in almost all cases the conflict resolves in the two final feet, where the ictus and the accent correspond. Once again, this has been well-studied in the traditional literature: consult in particular \citeA{knight_homodyne_1931} and \citeA[17-24]{duckworth_vergil_1969}. In the following discussion I adopt the terminology of these studies, and refer to a foot in which the ictus and accent coincide as \textit{homodyne}.%
\footnote{Knight and Duckworth both focus exclusively on conflict in the fourth foot. As will be seen later, conflict preferences in \textit{any} foot are important stylistic markers. Based on this analysis, conflict/harmony in the first foot (not the fourth) is actually the best place to look to distinguish different authors (see Table \ref{fig:feat_imp}).}
Finally, I measured \textit{elision}. When Latin words ending in vowels or nasals precede words that begin with a vowel, an elision occurs, and the final syllable of the first word is not counted for metrical purposes. Opinions differ as to whether elided syllables are completely dropped when the poem is read aloud, whether they are slurred or perhaps lightly pronounced---their pronunciation does not concern us here, only their presence. I measured the total number of elisions for each line (not including \textit{prodelision}, a slightly different feature)---the only non-binary feature.

While the metrical patterns and caesurae are trivial to extract from the MQDQ XML data, determining the ictus/accent conflicts required considerably more work. The general rule for word-accent in Latin is very simple, but there are a surprising number of exceptions and edge cases. In the end, it was necessary to develop new code to perform this analysis, which is incorporated into the open-source python package \texttt{MQDQParser} \cite{nagy_mqdq_2019}. Whenever new analysis tools are developed, care must be taken to ensure that the results are accurate---it is useless to base statistical analysis on bad data. There are at least two points of comparison. The first is Knight's `Homodyne in the fourth foot of the Vergilian Hexameter'. Knight informs us that ``[f]ourth-foot homodyne is by comparison rare in Vergil, but not very rare. The percentage of it to the total number of lines in the \textit{Aeneid} is 35.95 per cent" \cite[186]{knight_homodyne_1931}. Using my algorithm, fourth-foot homodyne was detected in 3507 lines (of 9840 in my edition), which is 35.65\%. The difference in our percentages represents perhaps 30 lines on which we disagree. Knight also lists some percentages by individual books. For \textit{Aen.} 1, he has 28.61\% while I have 28.82\%; \textit{Aen.} 8, 39.97\% to my 39.29\%; \textit{Aen.} 10, 32.59\% vs 32.71\%. While Knight does not share with us his methodology for scanning and tabulating the entire \textit{Aeneid} in the 1930s (perhaps it was a long winter), he does mention that he performed ``two computations, in which different texts were used and different conventions followed'' and that his figures varied by only 0.45\% \cite[186 n. 1]{knight_homodyne_1931}. The second source for independently determined statistics of this kind is \citeA[19-20, plus final Table]{duckworth_vergil_1969}. Duckworth consistently finds texts to have a slightly higher homodyne percentage than does Knight. Thus, for the entire \textit{Aeneid}, Duckworth records 37.78\% fourth-foot homodyne. This discrepancy is essentially due to disagreement among scholars regarding the correct accent of a few two syllable words (\textit{ergo, illuc,} \textellipsis), and whether certain conjunctions (\textit{et, at, aut, sed,} \textellipsis) bear any accent at all. Based on my research,
I have accepted the views of \citeA[88-9]{allen_vox_1965} (which apparently accord closely with Knight's), but this would be a poor venue in which to set out the entire debate.%
\footnote{For those that care to go beyond Allen, the best collection of the primary sources with respect to accent is \citeA{schoell_accentu_1876}, although the truly dedicated may prefer to read the works unabridged in \citeA{keil_heinrich_grammatici_1857}. The sources are, frankly, a mess of contradictions, and many of them are \nth{5} century or later, which stretches the appellation of `primary'.}
In any event, it can be hoped that as long as the analysis is being applied consistently to all texts that the statistical effect of any systemic errors will be negligible.

\subsection{Two-Label Classifiers}

Before attempting to use metrical features for any sort of authorship attribution, it should first be established that these features are able to distinguish authors at all. It would be foolish to rely on a measure that cannot do what an expert reader can do by instinct---to tell the difference between the hexameters of, say, Vergil and those of Ovid. To this end, I collected and analysed Vergil's \textit{Aeneid}; Ovid, \textit{Metamorphoses}; Statius, \textit{Thebaid}; Lucan, \textit{Pharsalia} and Silius' \textit{Punica}. These works are all composed in dactylic hexameters, and written as `epics'. Each pair was then analysed as a two-label classification problem with four algorithms: Extremely Random Trees, Gaussian Naïve Bayes, Logistic Regression and Support Vector Machines. All tests were carried out using the Python package \texttt{scikit-learn} \cite{scikit-learn}, and replication code is available in the github repository for this paper \cite{nagy_hexml_2019}.%
\footnote{The accompanying repository \cite{nagy_hexml_2019} also includes replication code for all Figures and Tables that appear in this paper.}
The tests were conducted with an 80/20 train-test split, with a chunk-size of 81 lines (to model the later authorship problem). In every case except Vergil vs Silius and Vergil vs Statius the accuracies were virtually 100\%. Because the most `difficult' classification is Silius vs Vergil, the rest of this section is mostly concerned with that pair. In Fig. \ref{fig:feat_effectiv_bw}, we can see that analsying foot patterns (\texttt{Feet Only}) is only slightly more effective than considering nothing but the patterns of ictus/accent conflict (\texttt{Conflict Only})---this underscores the value of conflict as an indicator of style. In fact, in terms of foot patterns, the importance of the first and fourth feet (dactyl or spondee) far outweighs the other two. It can be seen from Fig. \ref{fig:feat_effectiv_bw} that the set \texttt{Caes+Confl+FirstFoot} (Caesurae, Conflict and First Foot) performs almost as well as the entire feature set.

\begin{figure}
    \caption{Effectiveness of Different Feature Subsets}
    \label{fig:feat_effectiv_bw}
    \includegraphics[width=\textwidth]{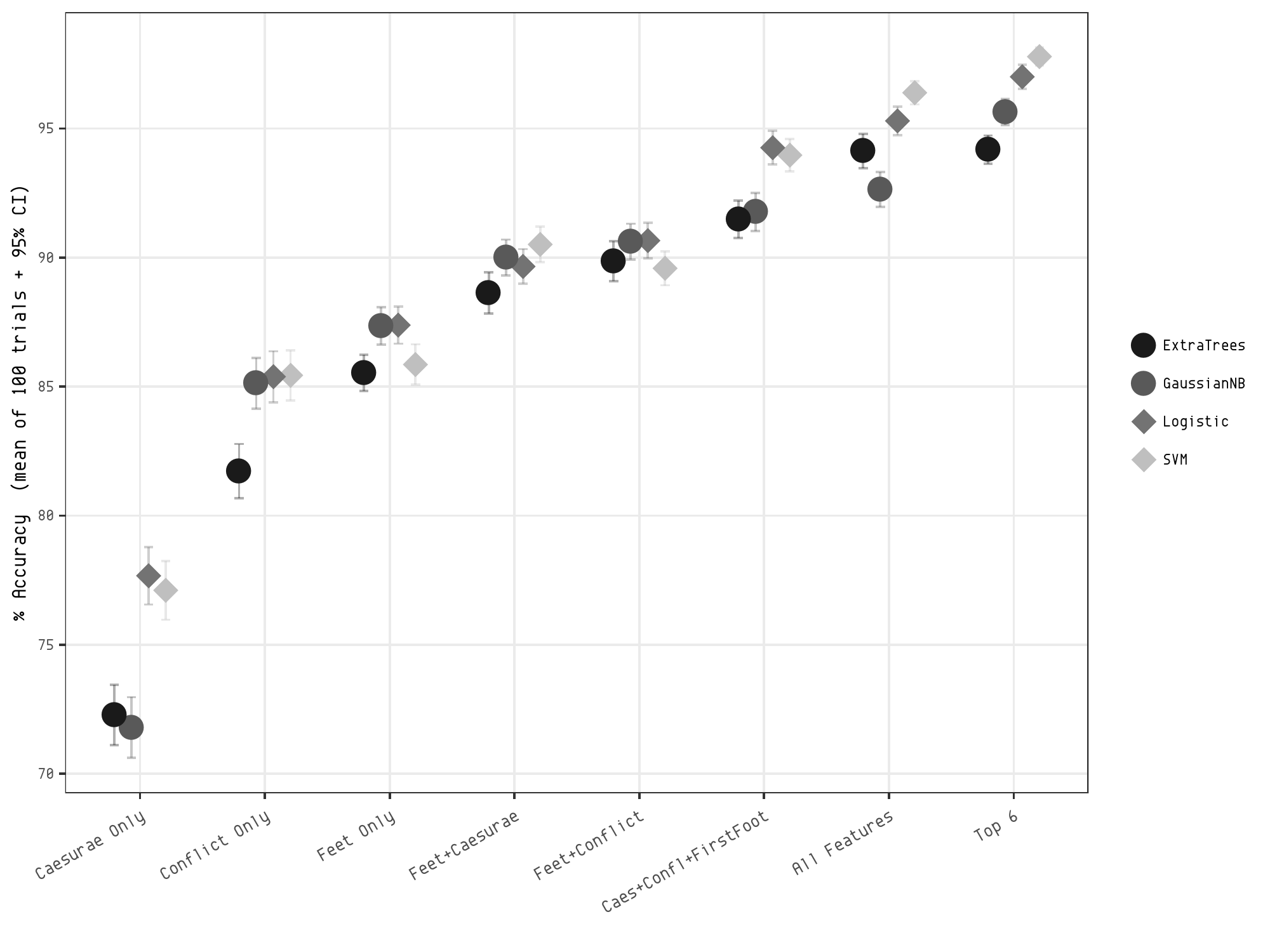}
\end{figure}

Another benefit of working with pairwise classification is that it can provide some kind of intuition as to which metrical features are the most useful to consider when distinguishing between authors. Not all machine learning algorithms can produce understandable feature importances, but it is possible to extract them from the Random Forest family of algorithms (this study used \texttt{ExtraTrees} from scikit-learn) and Support Vector Machines (with a linear kernel). I compared each pairwise combination of the five works above (10 total comparisons) and averaged the feature importances according to \texttt{SVM} and \texttt{ExtraTrees}, with the results shown in Table \ref{fig:feat_imp}. It is important to note that the two sets of values listed cannot be compared with each other, and it is not clear how the weights compare within a set (i.e. a feature with a weight of ten may not be ten times as important as one with a weight of one). In terms of raw ranking, though, they provide an interesting guide. 
\begin{table}
\caption{Ranked Feature Importances, as determined by \texttt{ExtraTrees} and \texttt{SVM}}
\label{fig:feat_imp}
\phantom{x}
\centering
\adjustbox{max width=\textwidth}{%
\begin{tabular}{ | l | c | l | c |}
\addlinespace
\hline
\multicolumn{2}{|c|}{\textbf{ExtraTrees}} & \multicolumn{2}{c|}{\textbf{SVM}}\\
\hline
Feature & Score & Feature & Score\\
\hline
\texttt{SYN} & 23.97 & \texttt{F1S} & 8.36\\
\texttt{F1S} & 18.02 & \texttt{SYN} & 7.67\\
\texttt{F4S} & 9.55 & \texttt{F3WC} & 5.29\\
\texttt{F3WC} & 7.14 & \texttt{F4S} & 4.25\\
\texttt{F1C} & 6.07 & \texttt{F4SC} & 3.99\\
\texttt{F3SC} & 5.07 & \texttt{F3C} & 3.92\\
\texttt{F3C} & 4.94 & \texttt{F3SC} & 2.79\\
\texttt{F3S} & 3.96 & \texttt{F2C} & 2.61\\
\texttt{F2C} & 3.58 & \texttt{F3S} & 2.54\\
\texttt{F4C} & 3.56 & \texttt{F4C} & 2.20\\
\texttt{F4SC} & 2.97 & \texttt{F2S} & 2.13\\
\texttt{F2WC} & 2.65 & \texttt{BD} & 2.08\\
\texttt{BD} & 2.11 & \texttt{F2WC} & 2.04\\
\texttt{F2S} & 1.87 & \texttt{F4WC} & 1.87\\
\texttt{F4WC} & 1.81 & \texttt{F1C} & 1.82\\
\texttt{F2SC} & 1.70 & \texttt{F2SC} & 1.72\\
\hline
\end{tabular}}
\end{table}
\subsection{Chunking}

After each line is converted to a feature vector, it can be considered as a single observation. However, because of the inherent variability of the lines, the two-label classifiers were all found to perform poorly on the raw data, with accuracies under 60\% (not much better than chance). This seems to be intuitively reasonable---a human reader would also find it more difficult to determine the author of a single line than to examine a hundred lines at once. The feature observations were, therefore, `chunked', by taking a set of lines and considering the set to be a single observation with the mean of the feature values; a set of twenty lines of which 15 had a strong third-foot caesura would have an \texttt{F3SC} value of 0.75. The trade-off, of course, is that the number of observations is reduced: 10,000 lines of the Aeneid with a chunk size of 20 becomes just 500 observations. The question then becomes `how many lines do we need in a chunk before we can make reliable decisions?'. The answer, of course, is `it depends'. Comparing Silius to Ovid, a 10-line chunk is enough to yield more than 90\% accuracy (see Fig. \ref{fig:clf_chunksz_silov_bw}). Silius and Vergil are the most similar,%
\footnote{Silius, it seems, admired the work of Vergil with almost religious fervour, cf Plin. \textit{Ep.} 3.7, 8.}
so, considering the performance of four different classification algorithms on this pair (Fig. \ref{fig:clf_chunksz_bw}), it seems reasonable to expect that anything over 60 lines should be enough to yield sensible results in general. It should also be noted that the order of the lines in the original works were randomised before being chunked. The intention of the randomisation process was to normalise `non-atmospheric' variation. Latin poets use the metre of their verses for poetic effect, and so Vergil will use different patterns in a battle scene than he does when describing funeral games---and yet (by hypothesis) his style is always `Vergilian'. By mixing the battles with the games, the love scenes with the death scenes, these surface variations are evened out, making it easier to see the deeper authorial style.

\begin{figure}
\centering
    \captionof{figure}{Classification Accuracy by Chunk Size: Silius vs Ovid}
    \label{fig:clf_chunksz_silov_bw}
    \includegraphics[width=\textwidth]{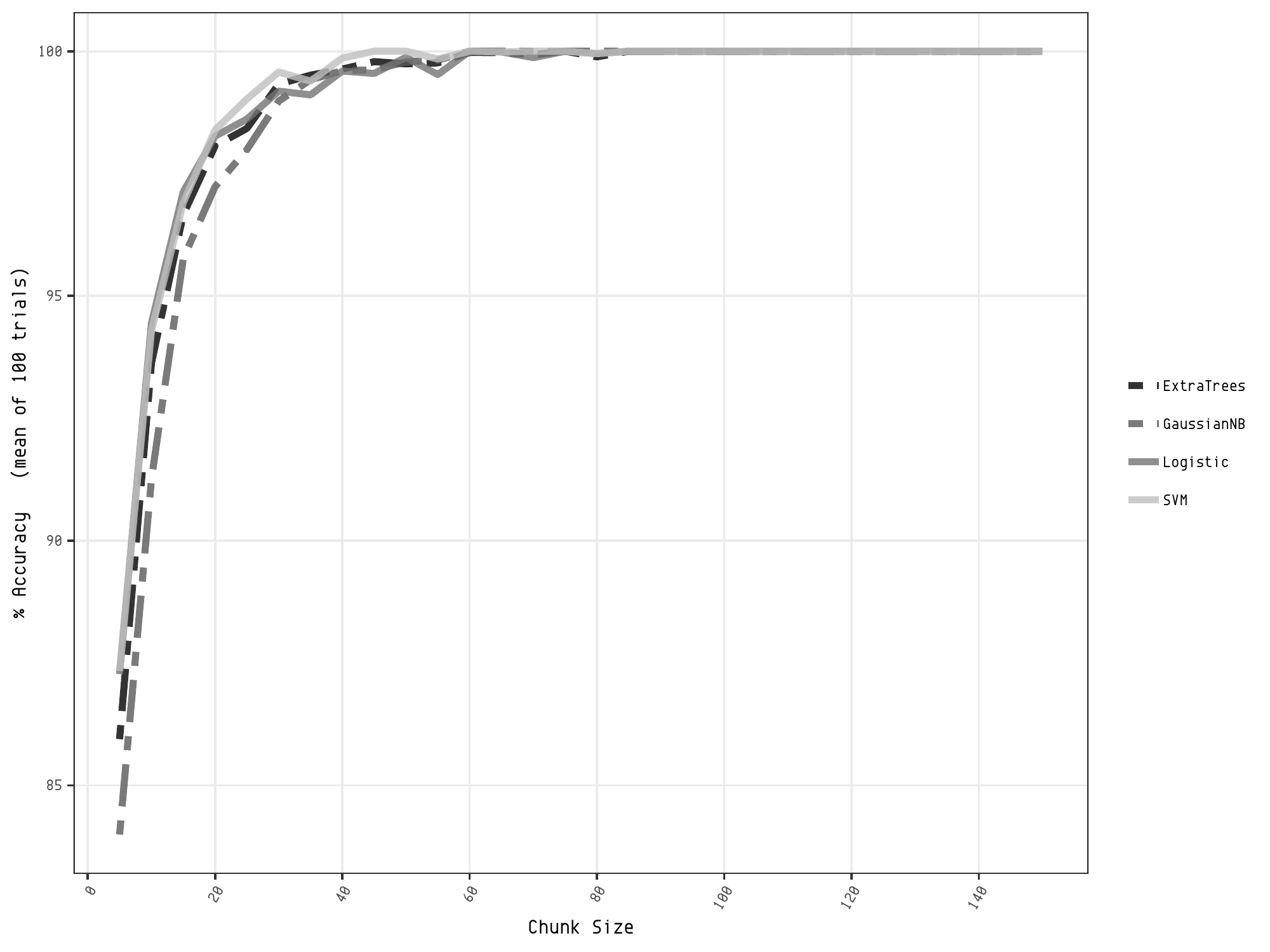}
\end{figure}
\begin{figure}
    \captionof{figure}{Classification Accuracy by Chunk Size: Silius vs Vergil}
    \label{fig:clf_chunksz_bw}
    \includegraphics[width=\textwidth]{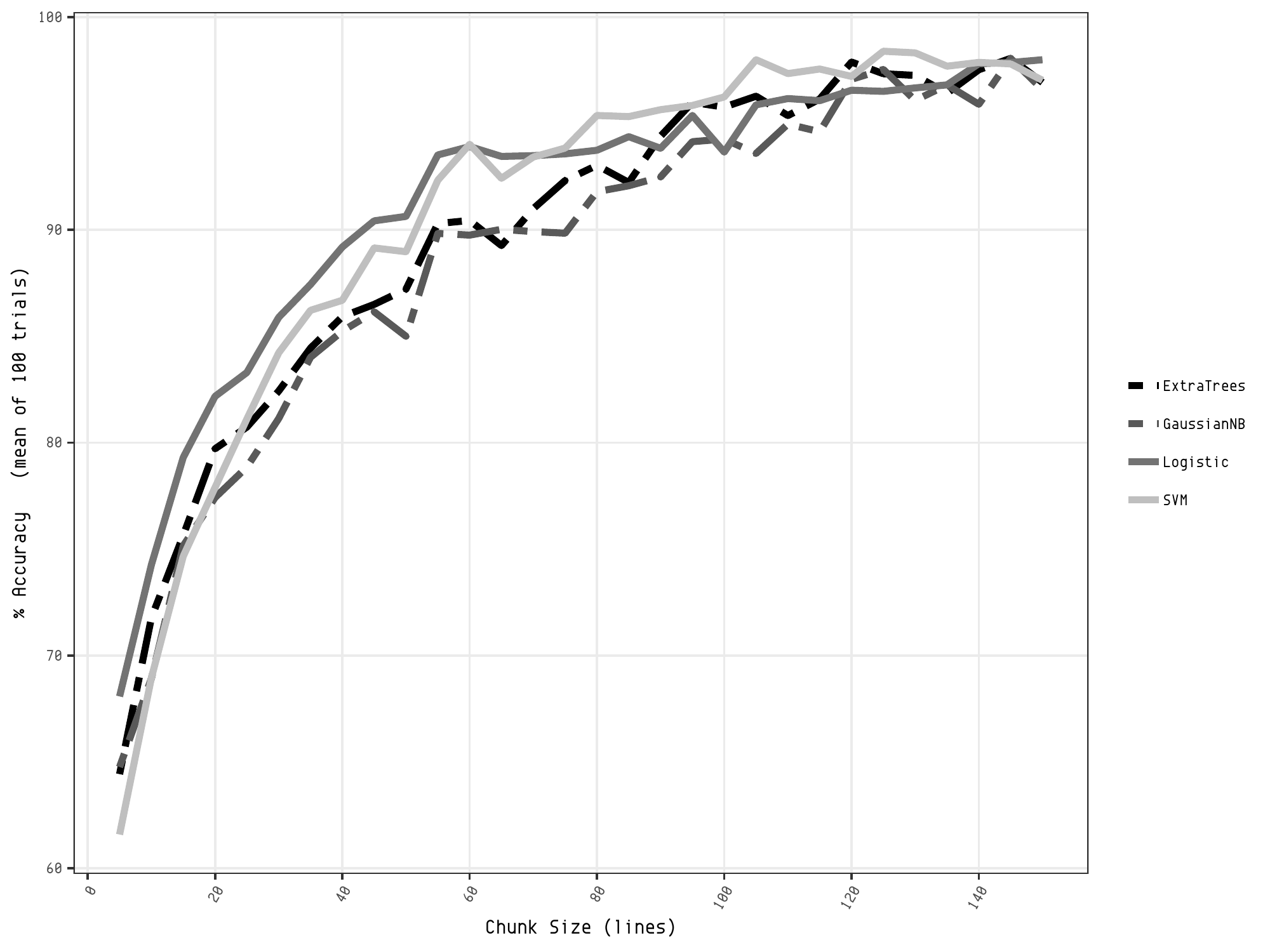}
\end{figure}

When considering the problem of reduced training sets, the low dimensionality of our feature set works to our advantage. Instead of considering hundreds of words or character $n$-grams, we have just 16 metrical features. When the Aldine Additamentum is taken as a single observation, and a chunk size of 81 lines is used for Silius' \textit{Punica} (the same length as the Additamentum) we have 150 observations---not a great number. For algorithms that are sensitive to dimensionality it would be better to have more than 256 (16$^{2}$), but the dimensionality does not seem unreasonable. If a smaller set of features is required, experiments indicate that standard feature selection techniques appear to work extremely well, with no loss of classification accuracy.%
\footnote{The downside, of course, is that the model will be less generally applicable.}
In Fig. \ref{fig:glob_feat_sel}, the best $n$ features were taken from the ordered list in Table \ref{fig:feat_imp} (as per SVM) and the Vergil vs Silius classification experiment was run with only those features. Of the global best features, the first eight perform as well or better as all 16. If the features are selected only with regard to Silius and Vergil, the accuracy is even better with six features than 16 (Fig. \ref{fig:feat_effectiv_bw}). This approach is not appropriate for problems where one author is completely unknown, but might be useful for a situation where there are a small number of candidate authors. Another possible approach to offset small training sizes due to chunking is to use repeated random sampling from the source lines. This is the approach taken in the following section.

\begin{figure}
    \caption{Parsimonious Models: Selecting a Feature Subset (results based on Silius vs Vergil, 81-line chunks)}
    \label{fig:glob_feat_sel}
    \includegraphics[width=\textwidth]{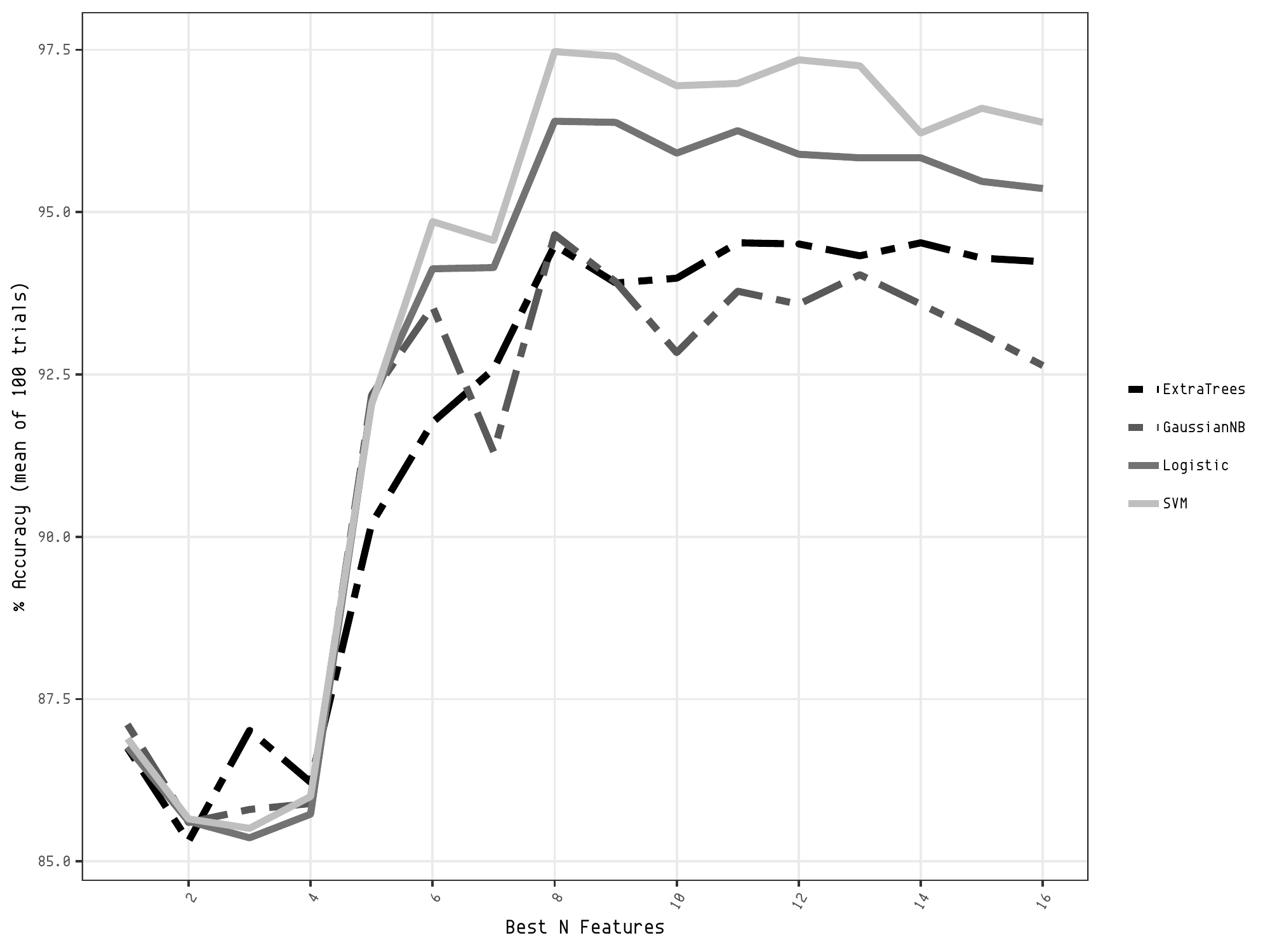}
\end{figure}

\subsection{Detecting `Scribal Interpolation' in Larger Works}

Having established that metrical features seem to be reliable indicators of authorial style, it now seems reasonable to apply them to an authorship problem. As mentioned, an interesting question is the authorship of the Aldine Additamentum.%
\footnote{For an excellent summary of background to this issue I refer the reader to \citeA[19-25]{lee_silius_2017}, whose thesis also includes a valiant attempt at univariate metrical analysis (26-8; Appendices 2-3, pp. 145-6), as well as a survey of the existing opinions in secondary literature (Appendix 1, p. 144).\label{leenote}}
At just over 500 words, the disputed text is too short to be a useful candidate for BOW or function-word approaches%
\footnote{As mentioned (above, n. \ref{edernote}), this question of ``how much text is enough?'' has been explored effectively by Eder (\citeyearNP{eder_does_2015} and \citeyearNP{eder_short_2017}).}
but, at 81 lines, it fits comfortably in the range where all of the two-author classification experiments achieved over 95\% accuracy. Broadly speaking, this problem fits best under the heading of `novelty detection'. There are a variety of appropriate algorithms that we could have selected, including \texttt{OneClassSVM} or \texttt{IsolationForest} but the simplest and most explanatory seemed to be a multi-dimensional cluster distance. The general approach, therefore, is this: we arrange the samples in a 16-dimensional feature space, choose a suitable distance metric and assume some kind of Gaussian distribution. Then, by calculating the distance of each point from the central tendency, we can assume the distances to be distributed according to $\chi^{2}$, and hence calculate a $p$-value. Because of the nature of the chunking, the features are already scaled in the range [0,1] (they are the mean of binary indicators), however as can be seen from Table \ref{fig:puni_corr}, there are some problems with co-variance and correlation.
\begin{table}
\caption{Some highly correlated features in Silius' \textit{Punica}}
\label{fig:puni_corr}
\phantom{x}
\centering
\adjustbox{max width=\textwidth}{%
\begin{tabular}{ | l | l | c |}
\hline
Feature 1 & Feature 2 & Pearson's $r$ \\
\hline\hline
\texttt{F3S} & \texttt{F3WC} & -0.502 \\
\texttt{F2C} & \texttt{F2WC} & -0.712 \\
\texttt{F3C} & \texttt{F3SC} & 0.787 \\
\texttt{F3C} & \texttt{F3WC} & -0.883 \\
\texttt{F4C} & \texttt{F4SC} & 0.869 \\
\texttt{F2SC} & \texttt{F2WC} & -0.512 \\
\texttt{F3SC} & \texttt{F3C} & 0.787 \\
\texttt{F3SC} & \texttt{F3WC} & -0.868 \\
\hline
\end{tabular}}
\end{table}
Accordingly, the most sensible distance was seen to be the Mahalanobis distance, which `normalises' the features by 'dividing' by the co-variance matrix. In mathematical terms, the Mahalanobis distance of an observation $x$ from the sample mean $\mu$ is defined to be $\sqrt{(x-\mu)^{T}C^{-1}(x-\mu)}$ (where $C$ is the covariance matrix of the target distribution $\textbf{X}$). In the following discussion, we will mainly see $M^{2}$, the square of the Mahalanobis distance, because it can be proven that $M^{2}$ follows the $\chi^{2}$ distribution, allowing for the direct calculation of a $p$-value. For this experiment, the target distribution was created by repeated random sampling.%
\footnote{An astute reader may wonder why, if repeated random sampling is a reasonable approach, it was not used during our classification experiments to offset the loss of sample size caused by line chunking. The problem is that if we had trained on a repeated random sample and then performed classification on `real' (sequential) chunks, much of the training data would re-appear (although in random places) in the test data. This would artificially inflate our accuracy and inevitably lead to over-fitting.}
The lines of the Additamentum were removed from the corpus, and then 10,000 sets of 81 lines were randomly selected with replacement. Based on experiment, this sample is large enough that there was negligible variation in $M^{2}$ due to the random process. The $M^{2}$ distance of the Additamentum from the target, 36.82, yields a corresponding $p$-value of 0.0013, low enough to be statistically significant. These results can be fully reproduced with the code in the accompanying github repository \cite{nagy_hexml_2019}. In an attempt to assess the reasonability of the final result, I made two other comparisons. When the Additamentum was compared to Vergil's \textit{Aeneid}, the corresponding $p$-value was 0.0006. Next, I compared the mean vector (or \textit{centroid}), representing an `average' chunk of Vergil, to Silius, and found a much lower distance of 15.88 with a $p$-value of 0.39 (recall, though, that we can tell Vergil from Silius with more than 95\% accuracy). In other words, the metrical style of the Additamentum differs much more from either Vergil or Silius than they differ from each other. These results are summarised in Table \ref{fig:addit_feat_comp}. On face value, these results support the theory that the Aldine Additamentum was not written by Silius---with the most likely alternative being that it was introduced by a humanist scholar in the late \nth{15} or early \nth{16} century to resolve a lengthy manuscript lacuna.%
\footnote{c.f Lee, (above, n. \ref{leenote})}

An additional advantage of the Mahalanobis distance%
\footnote{This may be a novel observation; certainly I have not come across it elsewhere in the literature.}
is that the distance can be easily decomposed to show a per-feature contribution to the total distance.%
\footnote{The core of the Mahalanobis distance is $(x-\mu)^{T}C^{-1}(x-\mu)$, and it is easy to see that this reduces to a vector dot product ($x^{T}y=k \in \mathbb{R}$). All that needs to be done is to multiply $xy$ point-wise and record the product vector. That vector is the feature contribution vector, and the sum of its entries is the squared Mahalanobis distance $M^{2}$.}
This \textit{feature contribution vector} seems to have good explanatory power, which can be extremely helpful in verifying results. For example, suppose that the contribution vector for a passage with a high $M^{2}$ distance indicates an unusual concentration of dactylic first feet. A poetic analysis might be able explain that divergence on stylistic grounds, considering the content of the passage. This kind of explanatory feature increases the \textit{transparency} of computational methods. The relevant feature contribution vectors for this section are shown in Table \ref{fig:addit_feat_comp}.

\begin{table}
\caption{Feature contribution vectors and Mahalanobis distance ($M^{2}$) when comparing the Additamentum to Silius' \textit{Punica} and Vergil's \textit{Aeneid}. For comparison, distance from the centroid of the \textit{Aeneid} to the \textit{Punica}.}
\label{fig:addit_feat_comp}
\phantom{x}
\centering
\adjustbox{max width=\textwidth}{%
\begin{tabular}{ | l | c | c | c |}
\hline
Feature & Addit. vs Punica & Addit. vs Aeneid & Aeneid vs Punica\\
\hline\hline
\texttt{F2WC} & 12.79 & 13.53 & 0.09\\
\texttt{F1C} & 7.43 & 3.48 & 0.21\\
\texttt{BD} & 7.06 & 6.26 & -0.03\\
\texttt{F4SC} & 2.88 & 0.18 & 1.80\\
\texttt{F2SC} & 2.31 & 3.45 & 0.16\\
\texttt{F3C} & 2.15 & 8.55 & 5.17\\
\texttt{F3SC} & 2.09 & 0.35 & 0.35\\
\texttt{F3WC} & 1.09 & 0.00 & 0.82\\
\texttt{F4S} & 0.83 & 0.90 & -0.00\\
\texttt{F4WC} & 0.46 & 1.10 & 0.02\\
\texttt{SYN} & 0.40 & 0.17 & 1.52\\
\texttt{F3S} & 0.29 & 0.04 & 0.05\\
\texttt{F1S} & 0.15 & 1.25 & 4.14\\
\texttt{F2S} & -0.21 & 0.64 & 0.21\\
\texttt{F4C} & -1.44 & 0.04 & -0.62\\
\texttt{F2C} & -1.45 & -0.59 & 2.00\\
\hline\hline
Raw Distance & 36.82 & 39.36 & 15.88\\
$p$-value & 0.0013 & 0.0006 & 0.39 \\
\hline
\end{tabular}}
\end{table}

\subsection{Interrogating the Authorship Results}

While these results are certainly very suggestive, it should be noted that they are not conclusive. Metre is only one aspect of poetic style, and it is certainly true that metre was consciously used by classical poets to impart an additional dimension to the verse---it is not an unbiased indicator. A worthwhile question to ask is whether some genuine passages might be this far from the `typical' style. As with any statistical determination, when considering enough samples this is certainly possible and, indeed, even likely. To illustrate this, I did the following: I first removed the Additamentum from the text of the Punica, and then took a rolling window of 81-line chunks throughout the text, advancing each time by 27 lines (so that the chunks overlap). At each step, the distance was calculated in the same way as for the Additamentum---removing that chunk from the rest of the poem, and then taking the $M^{2}$ distance based on a target distribution of 10,000 random 81-line samples. In plain terms, I tried to determine how many continuous sections of the undisputed text were at least as unusual as the Additamentum. These results are shown in Fig. \ref{fig:rolling_weirdness}. There were six chunks which were at least as unusual, or about 1.3\%. On examining them individually (albeit briefly), there seem to be plausible poetic reasons for the metrical distinctiveness of these chunks.%
\footnote{To explain this in full would require a lengthy discursion. However, as a brief example, the most extreme outlier, \textit{Puni.} 8.555--636, contains about double the average number of proper nouns, which require some metrical gymnastics to accommodate. Disregarding those lines brings the fragment back into the `fairly normal' zone. In addition, that section (surprisingly!) contains a number of short sections of deliberate rhyme which, once again, distort the statistics.}
Nevertheless, the results are enough to mandate caution. Although this evidence certainly supports the theory that the Additamentum is non-genuine, more work will be required to make the analysis conclusive. In particular, it seems sensible to attempt further tests using orthogonal feature sets, such as character $n$-grams, function words, rare lexicon or other approaches that have been similarly well-established in attribution studies.

\begin{figure}[h]
    \caption{Comparison: The Mahalanobis distance of undisputed lines, taking a rolling window of 81-line chunks. The Mahalanobis distance of the Additamentum is shown as a horizontal dotted line.}
    \label{fig:rolling_weirdness}
    \includegraphics[width=\textwidth]{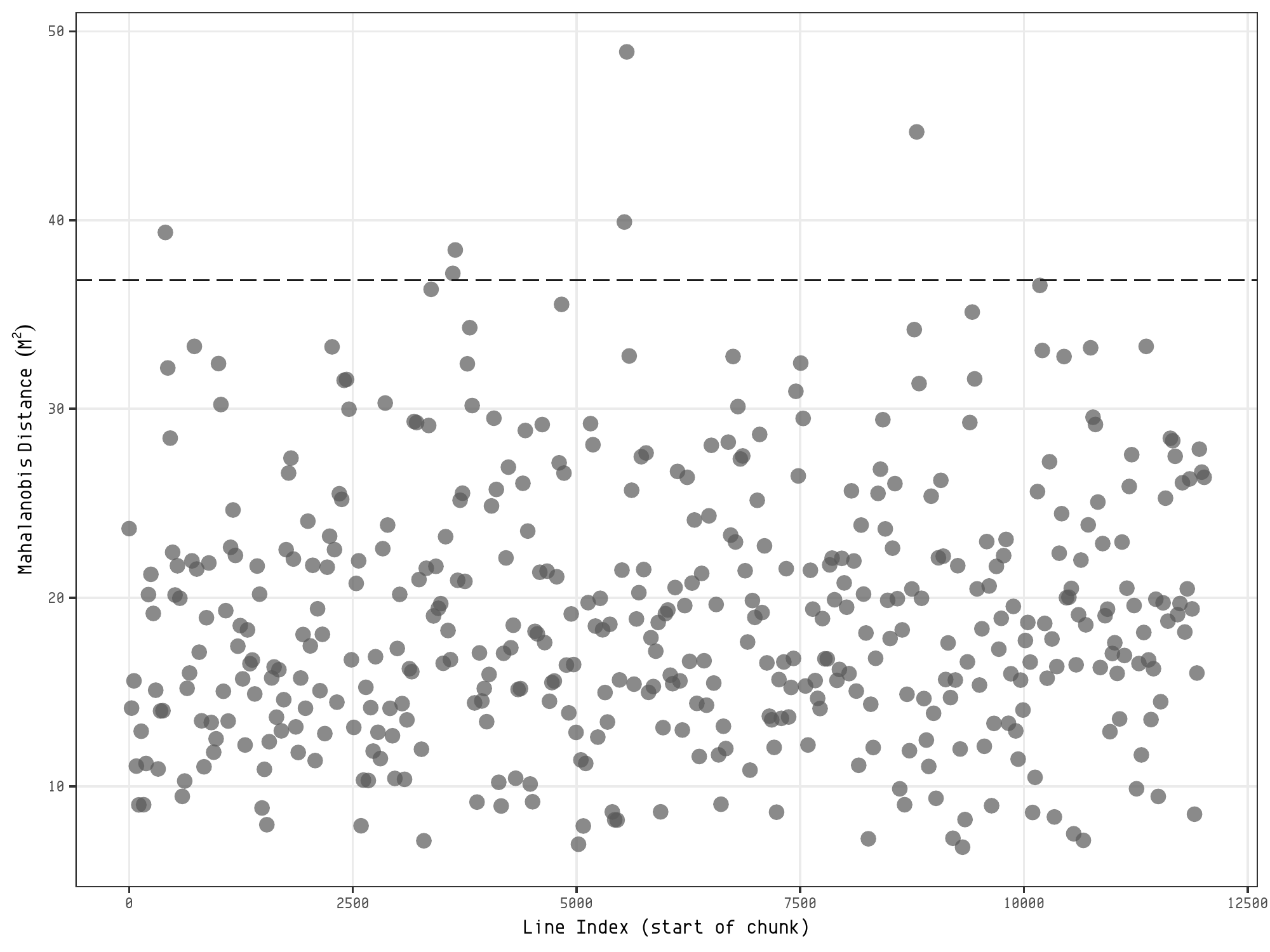}
\end{figure}

\section{Conclusion}
The clearest conclusion from this research is that metrical features are an extremely useful classifier for classical Latin hexameter poetry, and there seems little reason to doubt that they would be also effective for elegiac couplets. I note that this result is contra \citeA[293]{forstall_evidence_2011}, and must conclude that the difference lies in both the additional sophistication of the feature extraction (this study extracted more metrical features) and in the way the features were treated (it seems that $n$-gram frequencies on syllable quantity do not capture enough of the ambient stylistic information). In terms of further extensions, I am very optimistic about the prospects for the less common Latin metres, and would be reasonably surprised if the techniques were not also applicable to classical Greek. While single-line observations do not seem to be useful, chunked observations can return good results with chunks as small as ten lines in cases where styles are very different (Fig. \ref{fig:clf_chunksz_silov_bw}). Metrical features have low dimensionality, which may be useful in cases where limited training material is available. For Latin, a subset of eight features yielded very similar accuracy to all 16 features when used for classification. When designing experiments with metrical features, practitioners should bear in mind that some features are strongly correlated (Table \ref{fig:puni_corr}). In the classification tests, this did not seem to impede high accuracy.

In the authorship attribution test, the chosen metric was the Mahalanobis distance, which is designed to correct for the issues of correlation and co-variance. As was shown, the Mahalanobis distance has two other desirable features. First, it can be used directly to calculate a $p$-value, which provides a well understood difference indicator. Second, during the calculation of the Mahalanobis distance, it is possible to extract a \textit{feature contribution vector}, which seems to have useful explanatory properties (Table \ref{fig:addit_feat_comp})---this allows us to `bridge the gap' between opaque computational methods and traditional philology. To the best of my knowledge, this feature contribution vector has not been discussed in the stylometry literature. Based on this technique I analysed the disputed Aldine Additamentum in the \textit{Punica} by Silius Italicus. Here there were some important differences in methodology; the most noteworthy is that a target distribution was created through random sampling with replacement. These large sample distributions provided distance measurements that minimised variation due the random process. The results appear to support those scholars who claim that the work is not genuine. With a $p$-value of 0.0013, the Additamentum differs, metrically, from the style of the rest of the poem in a statistically significant way. For comparison, I examined the rest of the poem using a rolling window and found only six passages (of 446) with higher $M^{2}$ distances.%
\footnote{See Appendix, Table \ref{fig:weird_chunks} for a more detailed analysis of these passages.}
Although this authorship result is certainly promising, more research should be conducted using orthogonal methodologies before it could be considered to be a substantial claim.
\newpage
\bibliographystyle{apacite}
\bibliography{sil2}
\newpage
\section{Appendix}
\begin{table}[H]
\caption{Feature abbreviations used}
\label{fig:feat_abbrevs}
\phantom{x}
\centering
\begin{adjustbox}{width=\textwidth}
\begin{tabular}{ | l | l |}
\hline
Feat. & Description \\
\hline\hline
\texttt{F1S} & First Foot, Spondee (1 if Spondee, 0 if Dactyl)\\
\texttt{F2S} & Second Foot, Spondee\\
\texttt{F3S} & Third Foot, Spondee\\
\texttt{F4S} & Fourth Foot, Spondee\\
\texttt{F1C} & First Foot, Conflict (1 if there is an ictus/accent conflict, 0 for harmony)\\
\texttt{F2C} & Second Foot, Conflict\\
\texttt{F3C} & Third Foot, Conflict\\
\texttt{F4C} & Fourth Foot, Conflict\\
\texttt{BD} & Bucolic Diaeresis (1 if the fourth foot ends at a word break: \metricsymbols{_ u u ||} or \metricsymbols{_ _ ||})\\
\texttt{F2SC} & Second Foot, Strong Caesura (1 if there is a word break after the initial spondee of the second foot: \metricsymbols{_ ||})\\
\texttt{F3SC} & Third Foot, Strong Caesura\\
\texttt{F4SC} & Fourth Foot, Strong Caesura\\
\texttt{F2WC} & Second Foot, Weak Caesura (1 if there is a word break after the first breve. May only exist in dactylic feet: \metricsymbols{_ u ||})\\
\texttt{F3WC} & Third Foot, Weak Caesura\\
\texttt{F4WC} & Fourth Foot, Weak Caesura\\
\texttt{SYN} & Synalepha (elision) (A count of elisions, some lines may have more than one. Does not include prodelision.)\\
\hline
\end{tabular}
\end{adjustbox}
\end{table}

\begin{table}[H]
\caption{An analysis of the six outliers in Silius' \textit{Punica} (texts with $M^{2}$ higher than the Additamentum---see Fig. \ref{fig:rolling_weirdness}), for readers who might like to manually examine them.}
\label{fig:weird_chunks}
\phantom{x}
\centering
\adjustbox{max width=\textwidth}{%
\begin{tabular}{ | l | c | c | c || l | c | c | c || l | c | c | c |}
\hline
\textit{p} & $M^{2}$ & \multicolumn{2}{c||}{Book Ref.} & \textit{p} & $M^{2}$ & \multicolumn{2}{c||}{Book Ref.} & \textit{p} & $M^{2}$ & \multicolumn{2}{c|}{Book Ref.} \\
\hline

0.0007 & 38.92 & \multicolumn{2}{c||}{ 1:406--487} & 0.0014 & 36.66 & \multicolumn{2}{c||}{ 5:676--757} & 0.0008 & 38.43 & \multicolumn{2}{c|}{ 6:25--106} \\

\hline
Feat & Score & Samp\% & Mean\% & Feat & Score & Samp\% & Mean\% & Feat & Score & Samp\% & Mean\% \\
\hline

\texttt{F3WC} & 17.74 & 22.22 & 12.16 & \texttt{F2S} & 12.48 & 77.78 & 56.19 & \texttt{F2SC} & 17.43 & 85.19 & 62.64 \\
\texttt{F1S} & 5.69 & 64.20 & 50.66 & \texttt{F3WC} & 8.58 & 22.22 & 12.16 & \texttt{F3WC} & 14.41 & 23.46 & 12.16 \\
\texttt{F2C} & 4.64 & 72.84 & 78.34 & \texttt{F4S} & 7.97 & 59.26 & 72.66 & \texttt{F2S} & 10.02 & 75.31 & 56.19 \\
\texttt{F2WC} & 3.51 & 8.64 & 11.60 & \texttt{F2SC} & 4.09 & 77.78 & 62.64 & \texttt{F4C} & 3.12 & 67.90 & 60.67 \\
\texttt{F3SC} & 3.37 & 71.60 & 81.88 & \texttt{F4C} & 3.02 & 66.67 & 60.67 & \texttt{F4S} & 2.57 & 65.43 & 72.66 \\
\texttt{SYN} & 2.31 & 55.56 & 44.18 & \texttt{SYN} & 2.04 & 32.10 & 44.18 & \texttt{BD} & 1.19 & 55.56 & 50.88 \\
\texttt{BD} & 2.23 & 58.02 & 50.88 & \texttt{F2WC} & 1.38 & 2.47 & 11.60 & \texttt{F1C} & 0.96 & 50.62 & 42.52 \\
\texttt{F4C} & 1.96 & 65.43 & 60.67 & \texttt{F1C} & 0.52 & 48.15 & 42.52 & \texttt{F2C} & 0.95 & 86.42 & 78.34 \\
\texttt{F1C} & 1.77 & 50.62 & 42.52 & \texttt{BD} & 0.45 & 54.32 & 50.88 & \texttt{F4WC} & 0.25 & 3.70 & 4.89 \\
\texttt{F4WC} & 1.26 & 1.23 & 4.89 & \texttt{F3S} & 0.38 & 53.09 & 61.39 & \texttt{F1S} & -0.00 & 50.62 & 50.66 \\
\texttt{F2S} & 0.93 & 51.85 & 56.19 & \texttt{F1S} & 0.00 & 50.62 & 50.66 & \texttt{SYN} & -0.09 & 35.80 & 44.18 \\
\texttt{F3S} & 0.14 & 61.73 & 61.39 & \texttt{F4WC} & -0.03 & 4.94 & 4.89 & \texttt{F3S} & -0.46 & 54.32 & 61.39 \\
\texttt{F4S} & -0.04 & 74.07 & 72.66 & \texttt{F2C} & -0.56 & 86.42 & 78.34 & \texttt{F4SC} & -1.17 & 64.20 & 60.08 \\
\texttt{F2SC} & -0.33 & 64.20 & 62.64 & \texttt{F4SC} & -0.85 & 61.73 & 60.08 & \texttt{F3C} & -3.00 & 74.07 & 84.34 \\
\texttt{F4SC} & -0.42 & 61.73 & 60.08 & \texttt{F3SC} & -1.20 & 74.07 & 81.88 & \texttt{F2WC} & -3.39 & 3.70 & 11.60 \\
\texttt{F3C} & -5.85 & 79.01 & 84.34 & \texttt{F3C} & -1.61 & 75.31 & 84.34 & \texttt{F3SC} & -4.35 & 72.84 & 81.88 \\

\hline
\hline
\textit{p} & $M^{2}$ & \multicolumn{2}{c||}{Book Ref.} & \textit{p} & $M^{2}$ & \multicolumn{2}{c||}{Book Ref.} & \textit{p} & $M^{2}$ & \multicolumn{2}{c|}{Book Ref.} \\
\hline

0.0007 & 38.86 & \multicolumn{2}{c||}{ 8:528--609} & 0.0000 & 47.22 & \multicolumn{2}{c||}{ 8:555--636} & 0.0001 & 44.38 & \multicolumn{2}{c|}{13:442--523} \\

\hline
Feat & Score & Samp\% & Mean\% & Feat & Score & Samp\% & Mean\% & Feat & Score & Samp\% & Mean\% \\
\hline

\texttt{BD} & 12.03 & 33.33 & 51.02 & \texttt{F4SC} & 30.79 & 39.51 & 60.05 & \texttt{F2C} & 15.47 & 65.43 & 78.48 \\
\texttt{F3S} & 6.66 & 71.60 & 61.29 & \texttt{F3S} & 6.88 & 74.07 & 61.29 & \texttt{BD} & 7.32 & 34.57 & 51.02 \\
\texttt{F4SC} & 4.61 & 53.09 & 60.05 & \texttt{BD} & 6.62 & 39.51 & 51.02 & \texttt{F3SC} & 4.28 & 87.65 & 81.78 \\
\texttt{F3SC} & 3.42 & 74.07 & 81.78 & \texttt{F4S} & 3.43 & 80.25 & 72.65 & \texttt{F3WC} & 3.20 & 8.64 & 12.26 \\
\texttt{F3WC} & 3.25 & 18.52 & 12.26 & \texttt{F2WC} & 2.92 & 7.41 & 11.56 & \texttt{F4SC} & 2.58 & 69.14 & 60.05 \\
\texttt{F2WC} & 3.15 & 7.41 & 11.56 & \texttt{F4WC} & 2.47 & 9.88 & 4.87 & \texttt{F3S} & 2.40 & 69.14 & 61.29 \\
\texttt{F4WC} & 1.97 & 8.64 & 4.87 & \texttt{F3C} & 0.73 & 82.72 & 84.28 & \texttt{F3C} & 2.18 & 82.72 & 84.28 \\
\texttt{F4S} & 1.85 & 77.78 & 72.65 & \texttt{F1S} & 0.59 & 54.32 & 50.71 & \texttt{F2SC} & 1.80 & 70.37 & 62.74 \\
\texttt{F3C} & 1.19 & 77.78 & 84.28 & \texttt{F3WC} & 0.51 & 13.58 & 12.26 & \texttt{F1S} & 1.37 & 44.44 & 50.71 \\
\texttt{F2S} & 0.93 & 61.73 & 56.35 & \texttt{F3SC} & 0.11 & 82.72 & 81.78 & \texttt{F2S} & 1.22 & 50.62 & 56.35 \\
\texttt{F2C} & 0.36 & 77.78 & 78.48 & \texttt{F2SC} & 0.09 & 67.90 & 62.74 & \texttt{F4S} & 0.95 & 66.67 & 72.65 \\
\texttt{F1C} & 0.17 & 44.44 & 42.58 & \texttt{F1C} & 0.05 & 44.44 & 42.58 & \texttt{F4WC} & 0.92 & 7.41 & 4.87 \\
\texttt{F1S} & 0.09 & 51.85 & 50.71 & \texttt{F2S} & -0.01 & 56.79 & 56.35 & \texttt{F2WC} & 0.45 & 11.11 & 11.56 \\
\texttt{SYN} & 0.04 & 53.09 & 44.11 & \texttt{F2C} & -0.24 & 79.01 & 78.48 & \texttt{F4C} & 0.33 & 69.14 & 60.66 \\
\texttt{F4C} & -0.30 & 59.26 & 60.66 & \texttt{SYN} & -0.43 & 46.91 & 44.11 & \texttt{SYN} & 0.04 & 45.68 & 44.11 \\
\texttt{F2SC} & -0.56 & 65.43 & 62.74 & \texttt{F4C} & -7.28 & 50.62 & 60.66 & \texttt{F1C} & -0.14 & 40.74 & 42.58 \\
\hline
\end{tabular}}
\end{table}
\end{document}